\documentclass{article}
\usepackage{spconf,amsmath,graphicx}

\usepackage[latin1]{inputenc}

\usepackage{amsfonts}
\usepackage{amssymb}
\usepackage{scalerel}

\usepackage[T1]{fontenc}

\usepackage{amsthm}
\newtheorem{definition}{Definition}

\newcommand\scale[2]{\vstretch{#1}{\hstretch{#1}{#2}}}
\newcommand{\triangleplus}{\mathbin{\ooalign{$\bigtriangleup$\crcr\hidewidth
  \raise.14em\hbox{$\scale{0.7}{\scriptscriptstyle+}$}\hidewidth}}}
\newcommand{\triangleminus}{\mathbin{\ooalign{$\bigtriangleup$\crcr\hidewidth
  \raise.14em\hbox{$\scale{0.7}{\scriptscriptstyle-}$}\hidewidth}}}
\newcommand{\triangletimes}{\mathbin{  \ooalign{$\bigtriangleup$\crcr\hidewidth
  \raise.14em\hbox{$\scale{0.7}{\scriptscriptstyle\times}$}\hidewidth}}}

\newcommand{\Real}{\mathbb R}
\newcommand{\Tcurv}{\mathcal{T}}

\title{Aspl\"und's metric defined in the Logarithmic Image
Processing (LIP) framework for colour and multivariate images}
\twoauthors
  {Guillaume Noyel,	Member IEEE}
	{International Prevention Research Institute\\
	95 cours Lafayette\\
	69006 Lyon, France}
  {Michel Jourlin}
	{Laboratoire Hubert Curien, UMR CNRS 5516\\
	18 rue du Professeur Beno\^it Lauras\\
	42000 Saint-Etienne, France}

\begin{document}
\topmargin=0mm
%
\maketitle
\begin{abstract}
Aspl\"und's metric, which is useful for pattern matching, consists in a double-sided probing, i.e. the over-graph and the sub-graph of a function are probed jointly. It has previously been defined for grey-scale images using the Logarithmic Image Processing (LIP) framework. LIP is a non-linear model to perform operations between images while being consistent with the human visual system. Our contribution consists in extending the Aspl\"und's metric to colour and multivariate images using the LIP framework. Aspl\"und's metric is insensitive to lighting variations and we propose a colour variant which is robust to noise.
\end{abstract}

\section{Introduction}
\label{sec_intro}

The LIP framework has been originally defined by Jourlin et al. in \cite{Jourlin1988,Jourlin2001,Jourlin2011}. It consists in performing operations between images such as addition, subtraction or multiplication by a scalar with a result staying in the bounded domain of the images, for example $[0 ... 255]$ for grey-scale images on 8 bits. Due to the relation between LIP operations and the physical transmittance, the model is perfectly suited for images acquired by transmitted light (i.e. when the observed object is located between the source and the sensor). Furthermore, the demonstration, by Brailean \cite{Brailean1991} of the compatibility of the LIP model with human vision has enlarged its application for images acquired in reflected light.

The Aspl\"und's metric initially defined for binary shapes \cite{Asplund1960,Grunbaum1963} has been extended to grey-scale
images by Jourlin et al. using the LIP framework \cite{Jourlin2012,Jourlin2014}. One of the main property of this metric is to be strongly independent of lighting variations. It consists in probing a function by two homothetic functions of a template, i.e. the probe. As the homothetic functions are computed by a LIP multiplication, the distance is consistent with the human vision.

After a reminder of the LIP model and the Aspl\"und's metric for grey-scale images, a definition is given for colour images in order to perform colour matching. A variant of the metric, robust to the noise, is also proposed. Examples will illustrate the properties of the metric.

%
%

\section{Prerequisites}
\label{sec_pre}

%
%

\subsection{LIP model}
\label{sec_pre_LIP}

Given a spatial support $D \subset \mathbb{R}^{N}$, a grey-scale image is a function $f$ with values in the grey-scale $\left[0,M\right[ \in \mathbb{R}$:\\
$f: D \subset \Real^N \rightarrow \left[0,M\right[$.

In the LIP context, $0$ corresponds to the ``white'' extremity of the
grey-scale, which means to the source intensity, i.e. when no
obstacle (object) is placed between the source and the sensor.
Thanks to this grey-scale inversion, $0$ will appear as the neutral
element of the logarithmic addition.
The other extremity $M$ is a limit situation where no element of the
source is transmitted (black value).This value is excluded of the scale,
and when working with 8-bit digitized images, the 256 grey-levels
correspond to the interval of integers $\left[0... 255\right]$.

Due to the relation between the LIP model and the transmittance law, $T_f(x) = 1 - f(x)/M$ \cite{Jourlin2001},
the addition of two images $f$ and $g$ corresponds to the superposition of the obstacles (objects) generating respectively $f$ and $g$.
The resulting image will be noted:

$f \triangleplus g = f + g - \frac{f.g}{M}$

From this law, the multiplication of an image by a positive real number $\lambda$ is defined by:

$\lambda \triangletimes f = M - M \left( 1 - \frac{f}{M} \right)^{\lambda}$

Physical interpretation \cite{Jourlin2001}: In the case of transmitted light, the sum $2 \triangletimes f = f \triangleplus f$ consists in stacking twice the semi-transparent object corresponding to $f$.
Therefore, the LIP multiplication of $f$ by a scalar corresponds to changing the thickness of the observed object in the ratio $\lambda$. If $\lambda>1$, the thickness is increased and the image becomes darker than $f$, while if $\lambda<1$, the thickness is decreased and the image becomes brighter than $f$.

%
%

\subsection{Aspl\"und's metric for grey-scale images}
\label{sec_pre_Asp}

\begin{definition}
Given two images $f$ and $g$ defined on $D$, $g$ is chosen as a probing function for example, and we define the two
numbers: $\lambda = \inf \left\{\alpha, f \leq \alpha \triangletimes g \right\}$ and $\mu = \sup \left\{\beta, \beta \triangletimes g \leq f\right\}$. The corresponding ``functional Aspl\"und's metric'' $d_{As}^{\triangletimes}$ is:
\begin{equation}\label{eq:das}
	d_{As}^{\triangletimes}(f,g) = \ln \left( \lambda / \mu \right)
\end{equation}
\end{definition}

Physical interpretation \cite{Jourlin2014}: As Aspl\"und's metric is based on LIP multiplication, this metric is particularly insensitive to lighting variations, as long as such variations may be modelled by thickness changing.

In order to find in an image the location of a given template, the metric $d_{As}^{\triangletimes}$ can be adapted to local processing. The template corresponds to an image $t$ defined on a spatial support $D_t \subset D$. For each point $x$ of $D$, the distance $d_{As}^{\triangletimes}(f_{\left|D_t(x)\right.},t)$ is computed on the neighbourhood $D_t(x)$ centred in $x$, with $f_{\left|D_t(x)\right.}$ being the restriction of $f$ to $D_t(x)$.

%
%

\section{Aspl\"und's metric for colour and multivariate images}
\label{sec_asp}

A colour image $\mathbf{f}$, defined on a domain $D \subset \mathbb{R}^{N}$, with values in $\mathcal{T}^3$, $\mathcal{T} = \left[0,M\right[$, is written:
\begin{equation}\label{eq:f_col}
	\mathbf{f} : \left\{
	\begin{array}{ccc}
		D &\rightarrow& \mathcal{T}^3\\
		x &\rightarrow& \mathbf{f}(x) = \left( f_R(x) , f_G(x) , f_B(x) \right)\\
	\end{array}
	\right.
\end{equation}
with $f_R$ , $f_G$ , $f_B$ being the red, green and blue channels of $\mathbf{f}$, and $\mathbf{f}(x)$ being a vector-pixel.

A colour image is a particular case of a multivariate image which is defined as $\mathbf{f}_{\lambda} : D \rightarrow \mathcal{T}^L$, with $L$ being an integer number corresponding to the number of channels \cite{Noyel2007}. 

\begin{definition}
Given two colours $C_1 = (R_1,G_1,B_1)$, $C_2 = (R_2,G_2,B_2) \in \mathcal{T}^3$, their Aspl\"und's distance is equal to:
\begin{equation}\label{eq:d_As_C1C2}
 d_{As}^{\triangletimes} ( C_1 , C_2 ) =  \ln( \lambda / \mu )
\end{equation}
with $\lambda = \inf \left\{ k, k \triangletimes R_1 \geq R_2, k \triangletimes G_1 \geq G_2 ,  k \triangletimes B_1 \geq B_2 \right\}$
and  $\mu = \sup \left\{ k, k \triangletimes R_1 \leq R_2, k \triangletimes G_1 \leq G_2 ,  k \triangletimes B_1 \leq B_2 \right\}$
\end{definition}

Strictly speaking, $d_{As}^{\triangletimes}$ is a metric if the colours $C_n = (R_n, G_n, B_n)$ are replaced by their equivalence classes $\tilde{C}_n = \left\{ C=(R,G,B) \in \mathcal{T}^3 / \exists \alpha \in \Real^{+}, (\alpha \triangletimes R = R_n, \alpha \triangletimes G = \right.$ $\left. G_n, \alpha \triangletimes B = B_n )\right\}$.

Colour metrics between two colour images $\mathbf{f}$ and $\mathbf{g}$ may be defined as the sum ($d_1$ metric) or the supremum ($d_{\infty}$) of $d_{As}^{\triangletimes} ( C_1 , C_2 )$ on the considered region of interest $Z \subset D$:
\begin{align}
	d_{1,Z}^{\triangletimes}(\mathbf{f},\mathbf{g}) = \frac{1}{\#R} \sum_{x \in R} d_{As}^{\triangletimes} (\mathbf{f}(x),\mathbf{g}(x)) \\
	d_{\infty,Z}^{\triangletimes}(\mathbf{f},\mathbf{g}) = \sup_{x \in R} d_{As}^{\triangletimes} (\mathbf{f}(x),\mathbf{g}(x))
\end{align}
with $\#R$ the cardinal of $R$.
In the same way:

\begin{definition}
The global colour Aspl\"und's metric between two colour images $\mathbf{f}$ and $\mathbf{g}$ on a region $Z \subset D$ is
\begin{equation}\label{eq:d_As_C}
d_{As,Z}^{\triangletimes} (\mathbf{f},\mathbf{g}) = \ln( \lambda / \mu )
\end{equation}
with $\lambda = \inf \left\{ k, \forall x \in Z , k \triangletimes f_R(x) \geq g_R(x), k \triangletimes f_G(x) \geq \right.$ \\
	$\left. g_G(x) , k \triangletimes f_B(x) \geq g_B(x) \right\}$ \\
and $\mu      = \sup \left\{ k, \forall x \in Z , k \triangletimes f_R(x) \leq g_R(x), k \triangletimes f_G(x) \leq \right.$ \\
	$\left.  g_G(x) , k \triangletimes f_B(x) \leq g_B(x) \right\}$.
\end{definition}

\begin{figure}[!htb]

\begin{minipage}[b]{.46\linewidth}
  \centering
  \centerline{\includegraphics[width=\columnwidth]{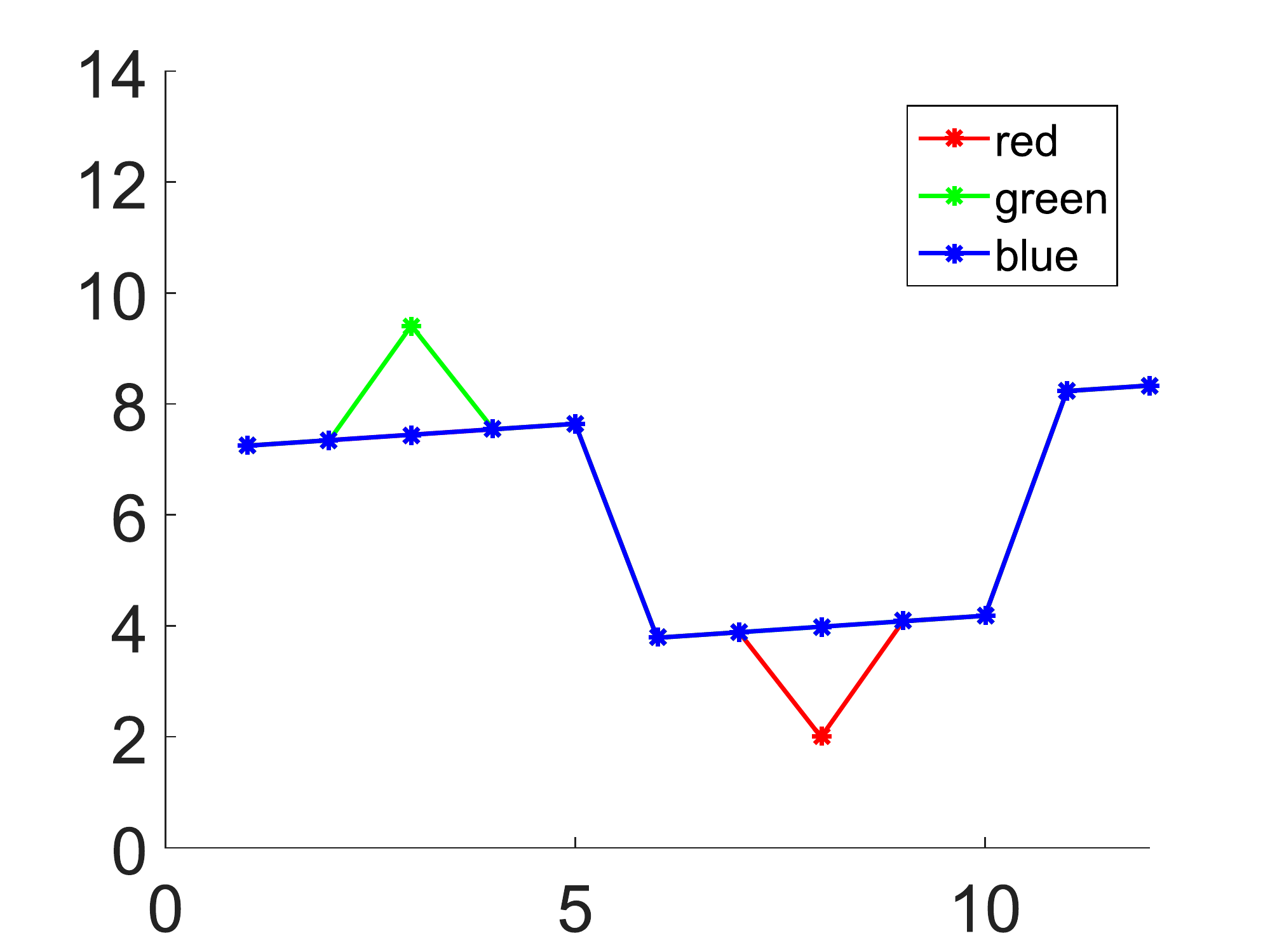}}
  \centerline{(a) Colour function $\mathbf{f}$}\medskip
\end{minipage}
\begin{minipage}[b]{.46\linewidth}
  \centering
  \centerline{\includegraphics[width=\columnwidth]{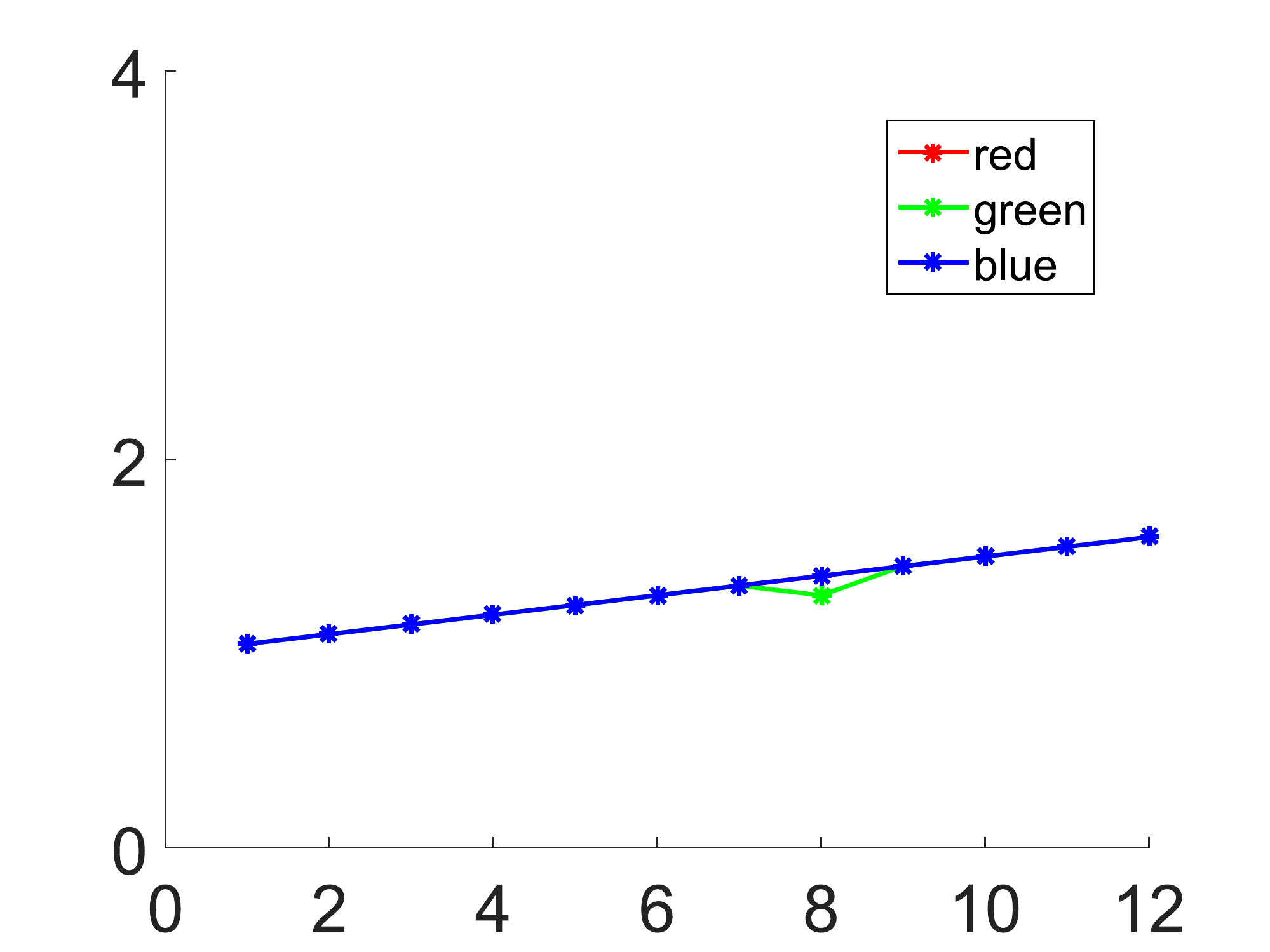}}
  \centerline{(b) Colour probe $\mathbf{g}$}\medskip
\end{minipage}
\hfill
\begin{minipage}[b]{0.46\linewidth}
  \centering
  \centerline{\includegraphics[width=\columnwidth]{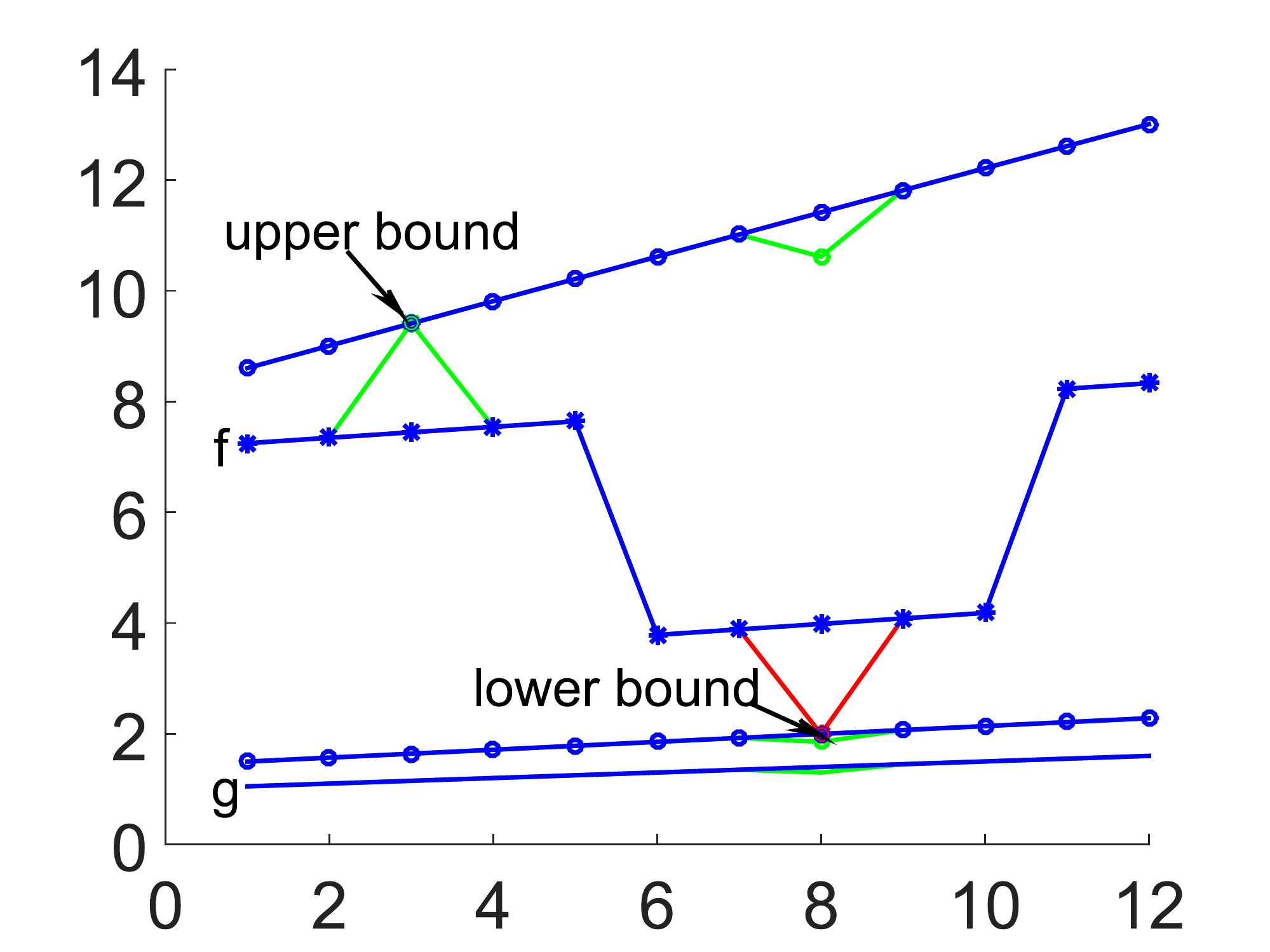}}
		(c) Lower ($\mu$) and upper ($\lambda$)	bounds
\end{minipage}

\caption{Computation of the Aspl\"und's distance between two colour functions $d_{As,D}^{\protect \triangletimes}(\mathbf{f},\mathbf{g}) = 1.76$. Each colour channel of the function is represented by a line having the corresponding colour.}
\label{fig:das_sig}
\end{figure}

In figure \ref{fig:das_sig}, the Aspl\"und's metric has been tested between the colour probe $\mathbf{g}$ and the colour function $\mathbf{f}$ on their definition domain $D$. In this case, the Aspl\"und's distance is equal to  $d_{As,D}^{\triangletimes}  (\mathbf{f},\mathbf{g}) = 1.76$. The lower bound $\mu$ and the upper bound $\lambda$ are determined as shown in Figure \ref{fig:das_sig} (c).

As for grey-scale images, the metric $d_{As,Z}^{\triangletimes}$ may be adapted to local processing. The template, (i.e. the probe) corresponds to a colour image $\mathbf{t}$ defined on a spatial support $D_t \subset D$. For each point $x$ of $D$, the distance $d_{As,D_t}^{\triangletimes} (\mathbf{f}_{\left|D_t(x)\right.},\mathbf{t})$ is computed on the neighbourhood $D_t(x)$ centred in $x$, with $\mathbf{f}_{\left|D_t(x)\right.}$ being the restriction of $\mathbf{f}$ to $D_t(x)$. 

\begin{definition}
Given a colour image $\mathbf{f}$ defined on $D$ into $\Tcurv^3$, $\left(\Tcurv^{3}\right)^{D}$, and a colour probe $\mathbf{t}$ defined on $D_t$ into $\Tcurv^{3}$, $\left(\Tcurv^{3}\right)^{D_{t}}$, the map of Aspl\"und's distances is:
\begin{equation}\label{eq:As_C}
	As_{\mathbf{t}}^{\triangletimes}\mathbf{f} : \left\{
	\begin{array}{ccc}
		\left(\Tcurv^{3}\right)^{D} \times \left(\Tcurv^{3}\right)^{D_{t}} &\rightarrow& {\Real^{+}}^{D}\\
		(\mathbf{f},\mathbf{t}) &\rightarrow& As_{\mathbf{t}}^{\triangletimes}\mathbf{f}(x) = \\
		&&d_{As,D_t}^{\triangletimes} (\mathbf{f}_{\left|D_t(x)\right.},\mathbf{t})\\
	\end{array}
	\right.
\end{equation}
with $D_t(x)$ the neighbourhood corresponding to $D_t$ centred in $x \in D$.
\end{definition}

\begin{figure}[!htb]
\begin{minipage}[b]{.49\linewidth}
  \centering
  \centerline{\includegraphics[width=\columnwidth]{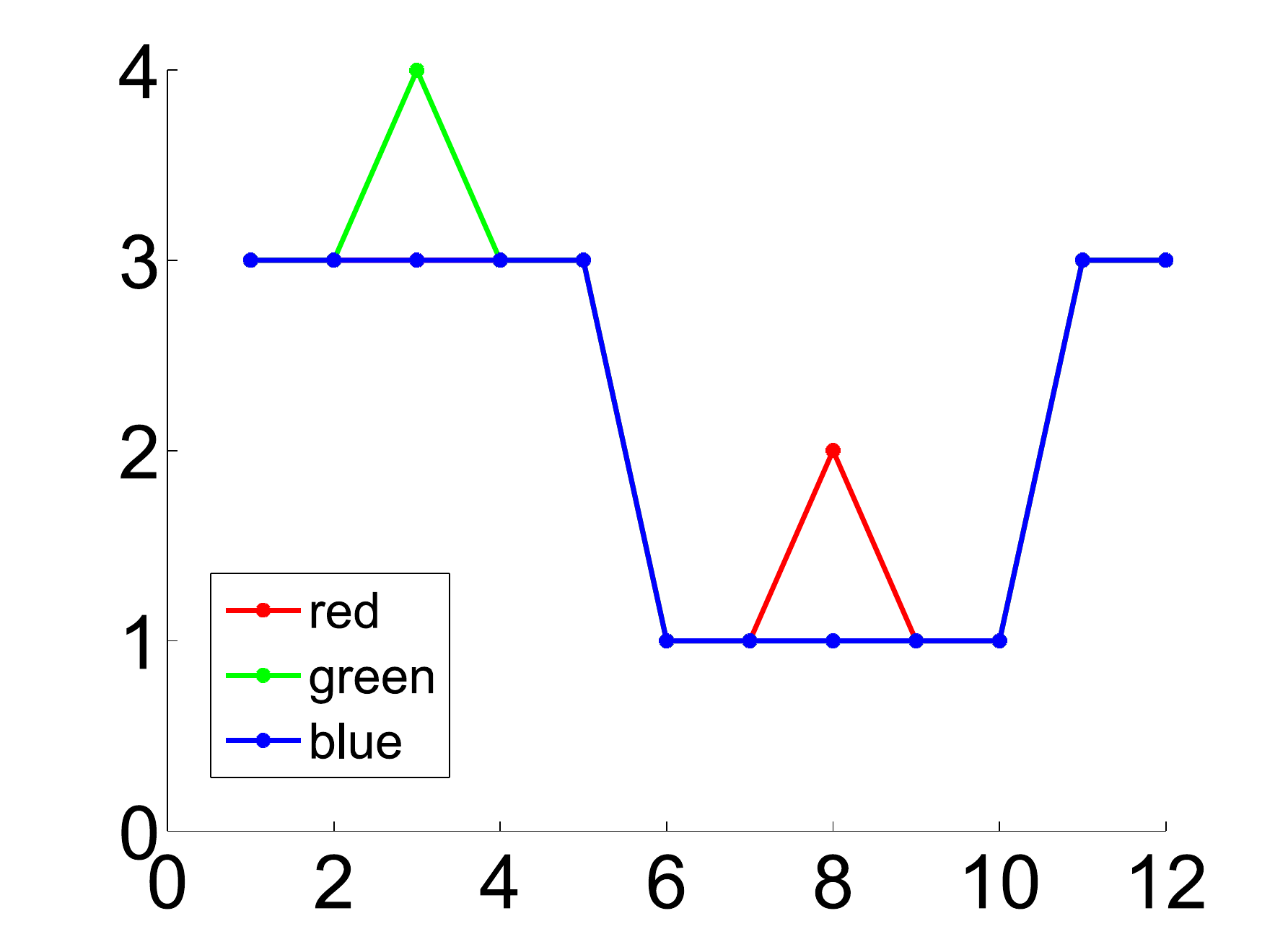}}
  \centerline{(a) Colour function $\mathbf{f}$}\medskip
\end{minipage}
\begin{minipage}[b]{.49\linewidth}
  \centering
  \centerline{\includegraphics[width=\columnwidth]{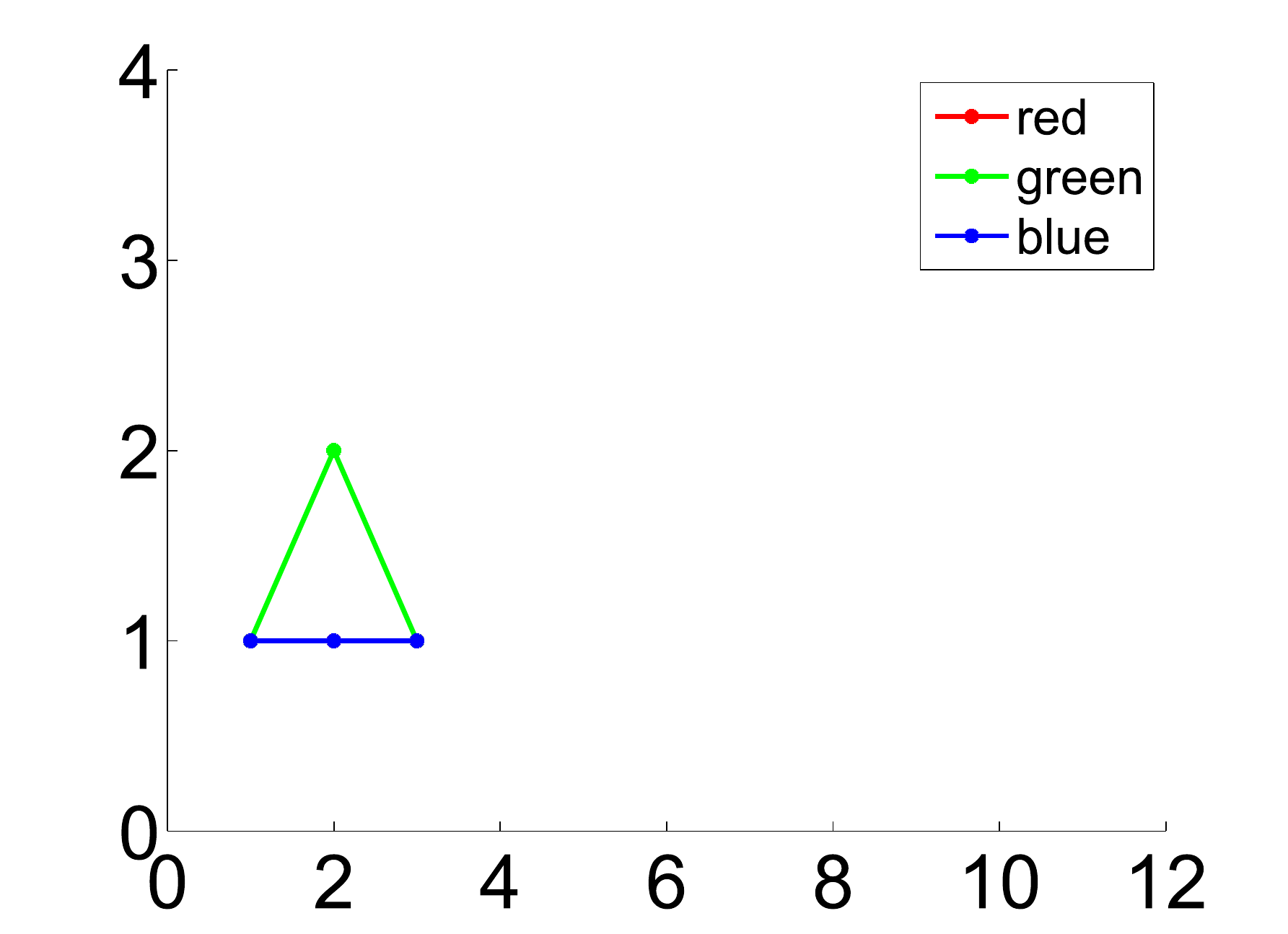}}
  \centerline{(b) Colour probe $\mathbf{t}$}\medskip
\end{minipage}
\hfill
\begin{minipage}[b]{0.49\linewidth}
  \centering
  \centerline{\includegraphics[width=\columnwidth]{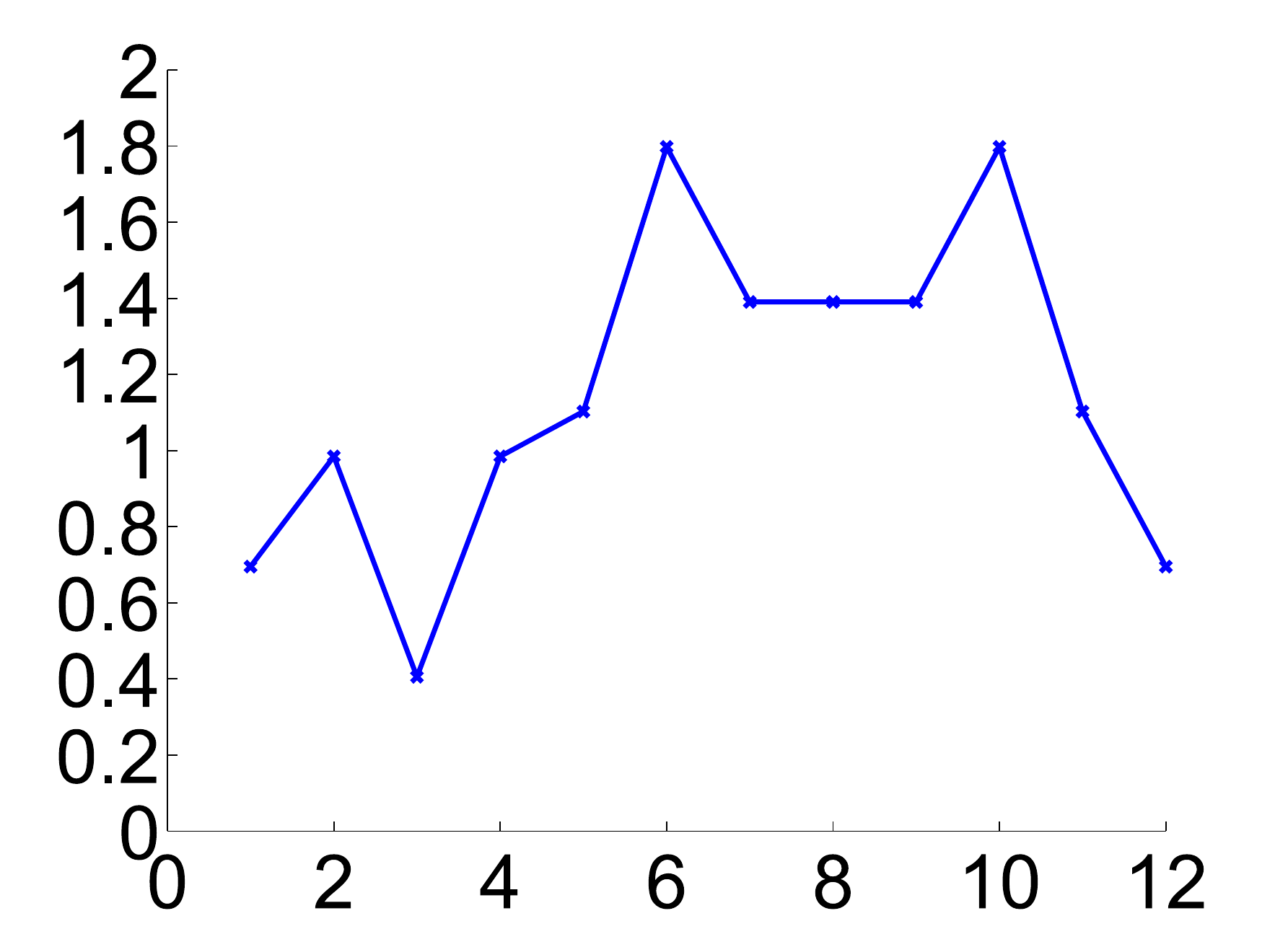}}
	(c) Map of Aspl\"und's\\
	distance $As_{\mathbf{t}}^{\triangletimes}\mathbf{f}$
\end{minipage}

\caption{(c) Map of the Aspl\"und's distance $As_{\mathbf{t}}^{\protect \triangletimes}\mathbf{f}$ between a colour function and a colour probe. (a) and (b) Each colour channel is represented by a line having the corresponding colour.}
\label{fig:map_as_sig}
\end{figure}

In figure \ref{fig:map_as_sig}, the map of Aspl\"und's distances is computed between a colour function and a colour probe. The minima of the map corresponds to the location of a pattern similar to the probe.

As explained in \cite{Jourlin2014}, Aspl\"und's distance is sensitive to noise because the probe lays on regional extrema produced by noise (Figure \ref{fig:das_sig}). In \cite{Jourlin2014}, Jourlin et al. have introduced a new definition of Aspl\"und's distance with a tolerance on the extrema corresponding to noise. In this paper, we extend this definition for colour and multivariate images.

To reduce the sensitivity of Aspl\"und's distance to the noise, there exists a
metric defined in the context of ``Measure Theory''. It will be called Measure metric or M-metric. Only a short recall of this theory adapted
to the context of colour images defined on a subset $D \subset \mathbb{R}^N$ is presented.
Given a measure $\mu$ on $\mathbb{R}^N$, a colour image $\mathbf{f}$ and a metric $d$ on the
space of  colour images, a neighbourhood $N_{\mu,d,\epsilon,\epsilon'}$ of function $\mathbf{f}$ may be
defined thanks to $\mu$ and two arbitrary small positive real
numbers $\epsilon$ and $\epsilon'$ according to:

	\[N_{\mu,d,\epsilon,\epsilon'} = \left\{ \mathbf{g}, \mu( x \in D, d(\mathbf{f}(x), \mathbf{g}(x)) > \epsilon ) < \epsilon' \right\}\]

It means that the measure of the set of points $x$, where $d(\mathbf{f}(x), \mathbf{g}(x))$ exceeds the tolerance $\epsilon$, satisfies another tolerance $\epsilon'$. This can be simplified, in the context of Aspl\"und's metric:

\begin{itemize}

	\item the image being digitized, the number of pixels lying in $D$ is finite, therefore the ``measure'' of a subset of $D$ is linked to the cardinal of this subset, for example the percentage $P$ of its elements related to $D$ (or a region of interest $R\subset D$). In this case, we are looking for a subset $D'$ of $D$, such that $\mathbf{f}_{\left|D'\right.}$ and $\mathbf{g}_{\left|D'\right.}$ are neighbours for Aspl\"und's metric and the complementary set $D\setminus D'$ of $D'$ related to $D$ is small sized when compared to $D$. This last condition can be written as:\\
 $P(D \setminus D') = \frac{\#(D \setminus D')}{\#D} \leq p$\\
where $p$ represents an acceptable percentage and $\#D$ the number of elements in $D$.

	\item Therefore, the neighbourhood $N_{\mu,d,\epsilon,\epsilon'}(f)$ is
	\begin{equation}
	\begin{array}{l}
		N_{P,d_{As},\epsilon,p}(\mathbf{f}) = \left\{ \mathbf{g} \setminus \exists D' \subset D, d_{As,D'}^{\triangletimes}(\mathbf{f}_{\left|D'\right.},\mathbf{g}_{\left|D'\right.}) < \epsilon \right.\\
		\left. \text{ and } \frac{\#(D \setminus D')}{\#D} \leq p\right\}
		\end{array}
	\end{equation}
	
\end{itemize}

\begin{figure}[!htb]
\begin{minipage}[b]{.49\linewidth}
  \centering
  \centerline{\includegraphics[width=\columnwidth]{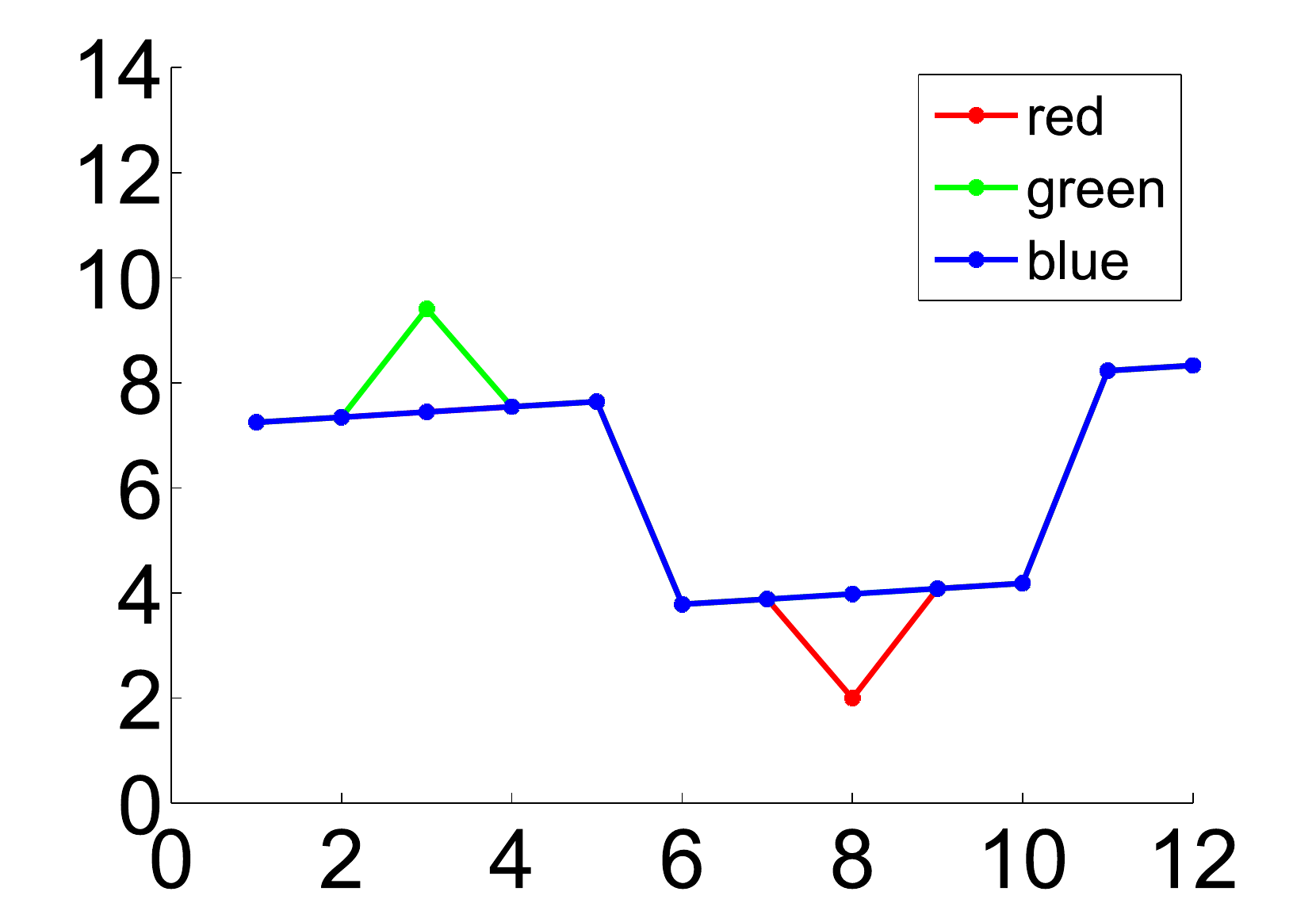}}
  \centerline{(a) Colour function $\mathbf{f}$}\medskip
\end{minipage}
\begin{minipage}[b]{.49\linewidth}
  \centering
  \centerline{\includegraphics[width=\columnwidth]{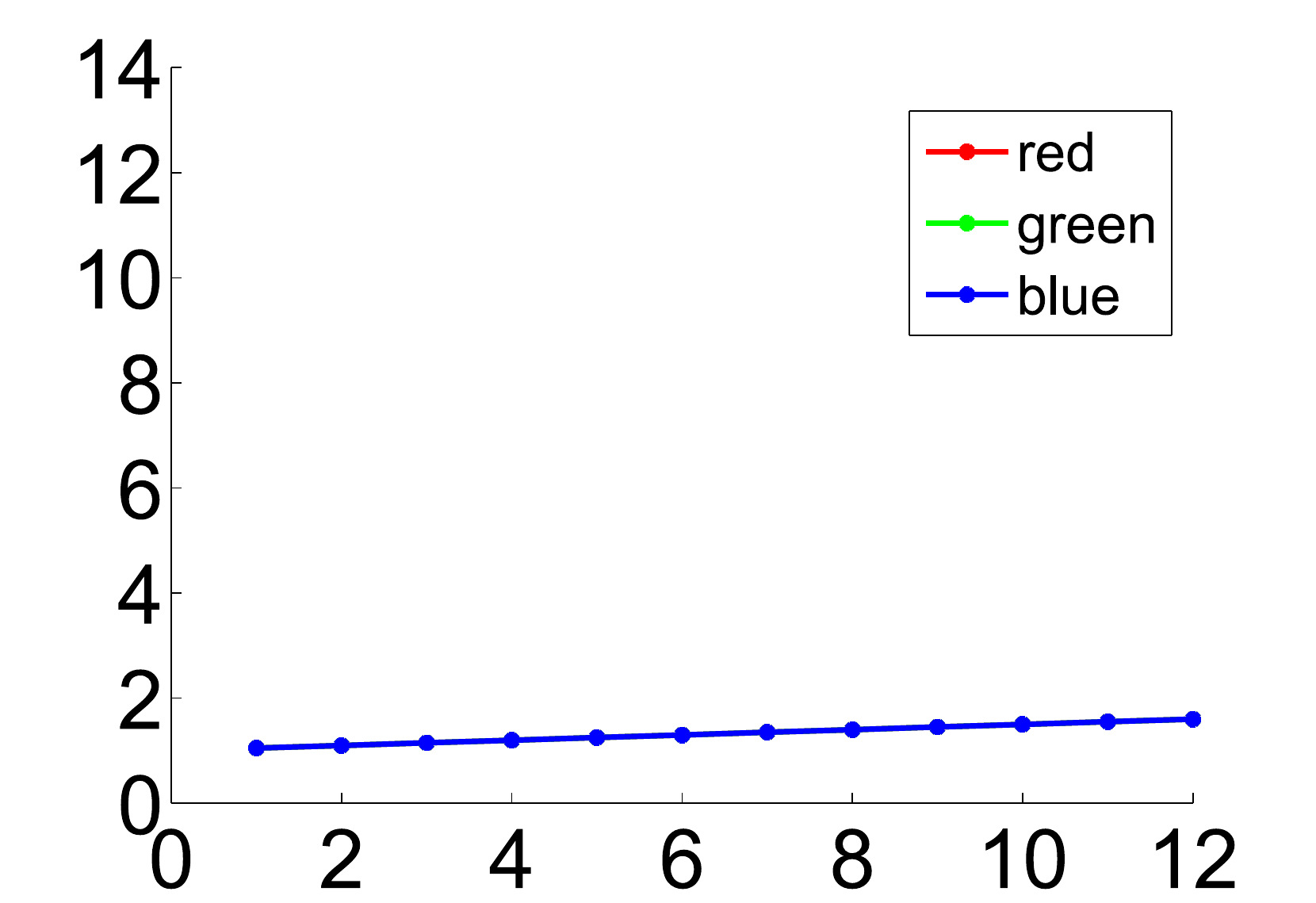}}
  \centerline{(b) Colour probe $\mathbf{g}$}\medskip
\end{minipage}
\hfill
\begin{minipage}[b]{0.67\linewidth}
  \centering
  \centerline{\includegraphics[width=\columnwidth]{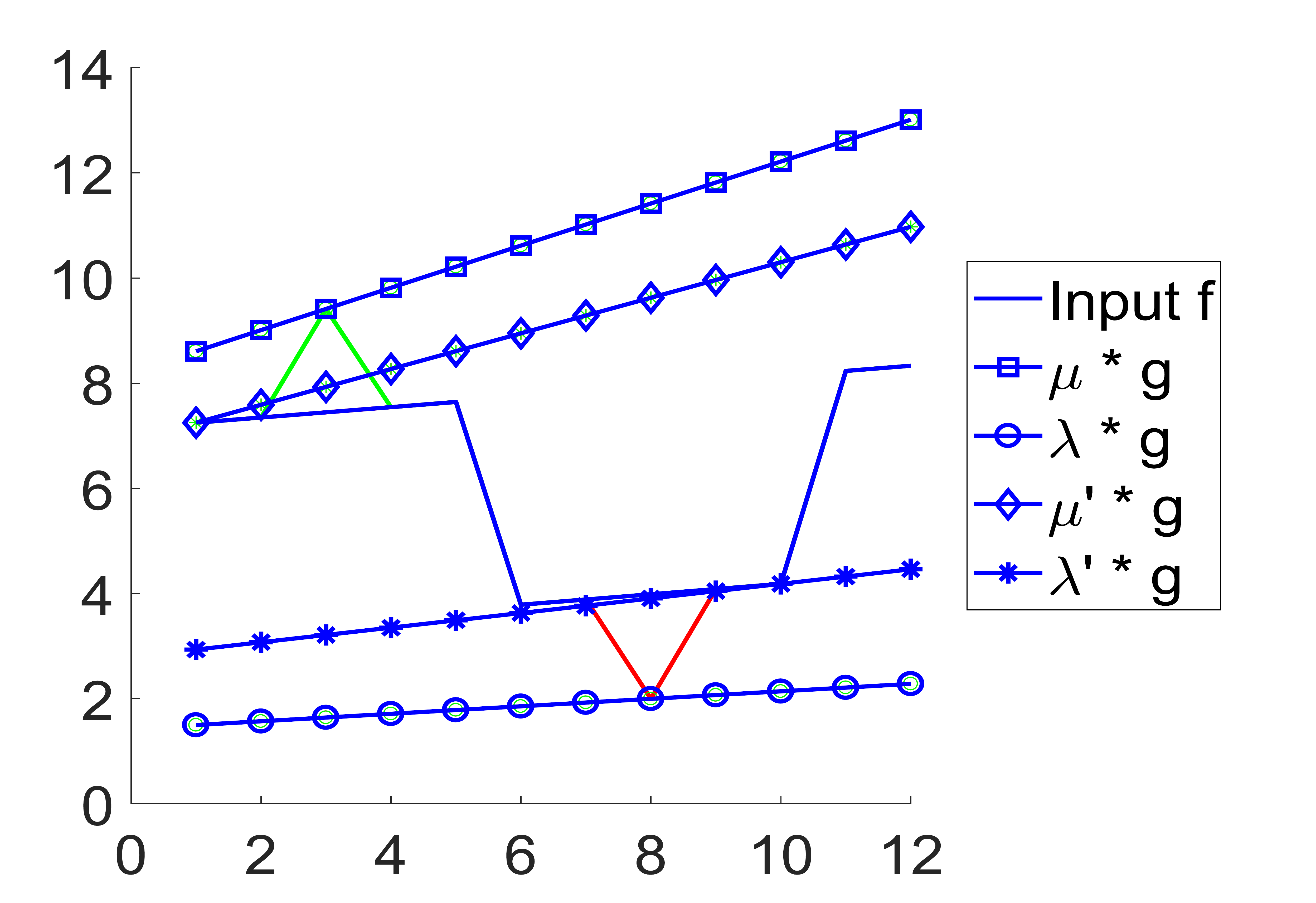}}
  \centerline{(c) Lower and upper bounds, $p = 80\%$ }
\end{minipage}
\caption{Colour Aspl\"und's distance with a tolerance of $p = 80 \%$. ($\mu$, $\lambda$) are the scalars multiplying the probe without tolerance. ($\mu'$, $\lambda'$) are the scalars multiplying the probe with tolerance.}
\label{fig:d_as_tol}
\end{figure}

We follow the same approach already used in \cite{Jourlin2014} to compute the Aspl\"und's distance with a tolerance $d_{As,D,p=80\%}^{\triangletimes}(\mathbf{f},\mathbf{g})$.
In figure \ref{fig:d_as_tol}, a tolerance of $p = 80\%$ is used and consists in discarding two points. Therefore, the Aspl\"und's distance is decreased from $d_{As,D}^{\triangletimes}(\mathbf{f},\mathbf{g}) = 1.76$ to $d_{As,D,p=80\%}^{\triangletimes}(\mathbf{f},\mathbf{g}) = 0.91$.

With this distance, a map of Aspl\"und's distances can be defined.
\begin{definition}
Given a colour image $\mathbf{f}$ defined on $D$ into $\Tcurv^3$, $\left(\Tcurv^{3}\right)^{D}$, and a colour probe $\mathbf{t}$ defined on $D_t$ into $\Tcurv^{3}$, $\left(\Tcurv^{3}\right)^{D_{t}}$, a tolerance $p\in[0,1]$, the map of Aspl\"und's distances with a tolerance is:
\begin{equation}\label{eq:As_C_tol}
	As_{\mathbf{t},p}^{\triangletimes}\mathbf{f} : \left\{
	\begin{array}{ccc}
		\left(\Tcurv^{3}\right)^{D} \times \left(\Tcurv^{3}\right)^{D_{t}} &\rightarrow& {\Real^{+}}^{D}\\
		(\mathbf{f},\mathbf{t}) &\rightarrow& As_{\mathbf{t},p}^{\triangletimes}\mathbf{f}(x) = \\
		& & d_{As,D_t,p}^{\triangletimes} (\mathbf{f}_{\left|D_t(x)\right.},\mathbf{t})
	\end{array}
	\right.
\end{equation}
with $D_t(x)$ the neighbourhood corresponding to $D_t$ centred in $x \in D$.
\end{definition}

After having introduced the colour Aspl\"und's distance, examples are given.

%
%

\section{Examples and applications}
\label{sec_ex}

\begin{figure}[!htb]
\begin{minipage}[b]{0.49\linewidth}
  \centering
  \centerline{\includegraphics[width=0.98\columnwidth]{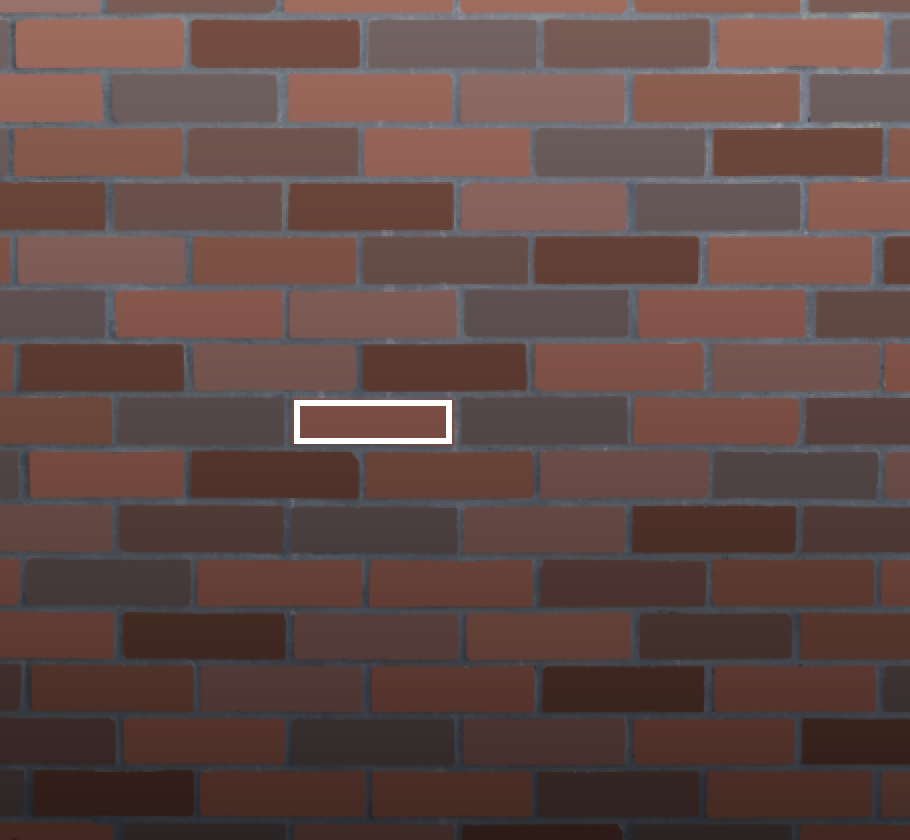}}
  \centerline{(a) Image $\mathbf{f}$ and probe $\mathbf{t}$}\medskip
\end{minipage}
\begin{minipage}[b]{0.49\linewidth}
  \centering
  \centerline{\includegraphics[width=0.98\columnwidth]{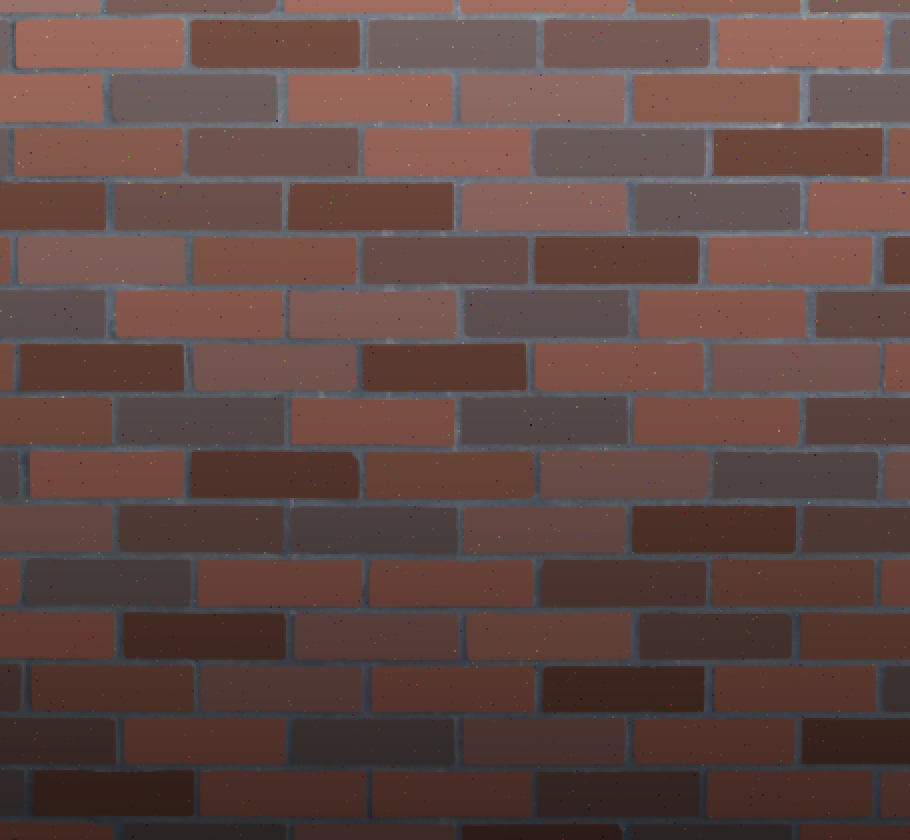}}
  \centerline{(b) Noisy image $\tilde{\mathbf{f}}$}\medskip
\end{minipage}
\hfill
\begin{minipage}[b]{0.49\linewidth}
  \centering
  \centerline{\includegraphics[width=0.98\columnwidth]{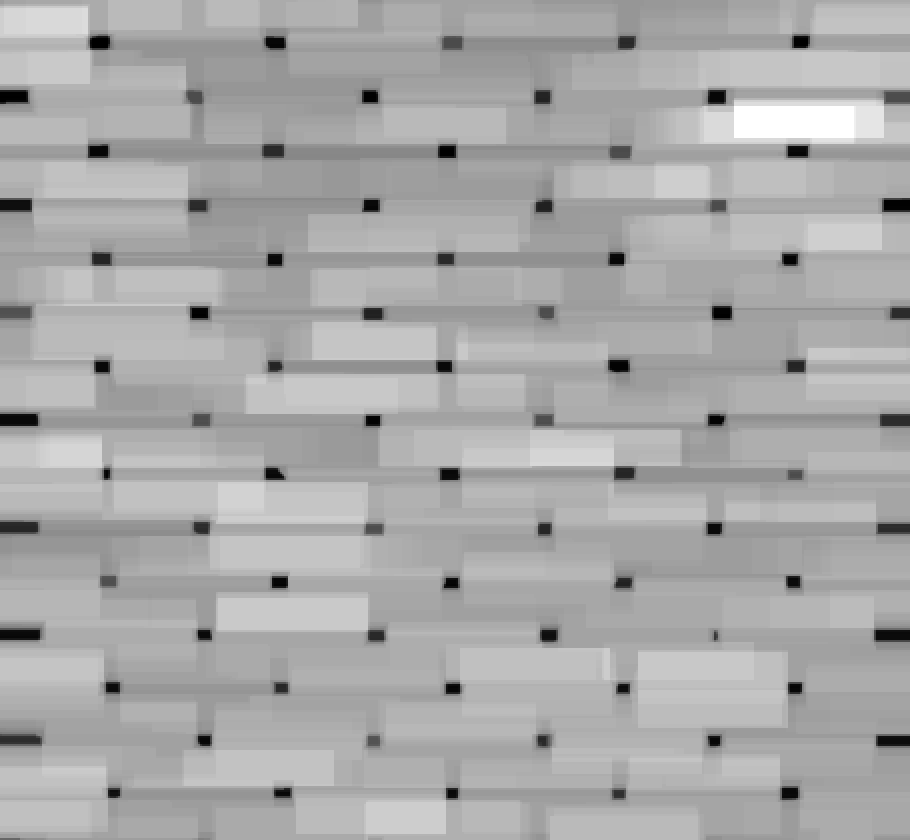}}
  \centerline{(c) Map $As_{\mathbf{t}}^{\protect \triangletimes}\mathbf{f}$}\medskip
\end{minipage}
\begin{minipage}[b]{0.49\linewidth}
  \centering
  \centerline{\includegraphics[width=0.98\columnwidth]{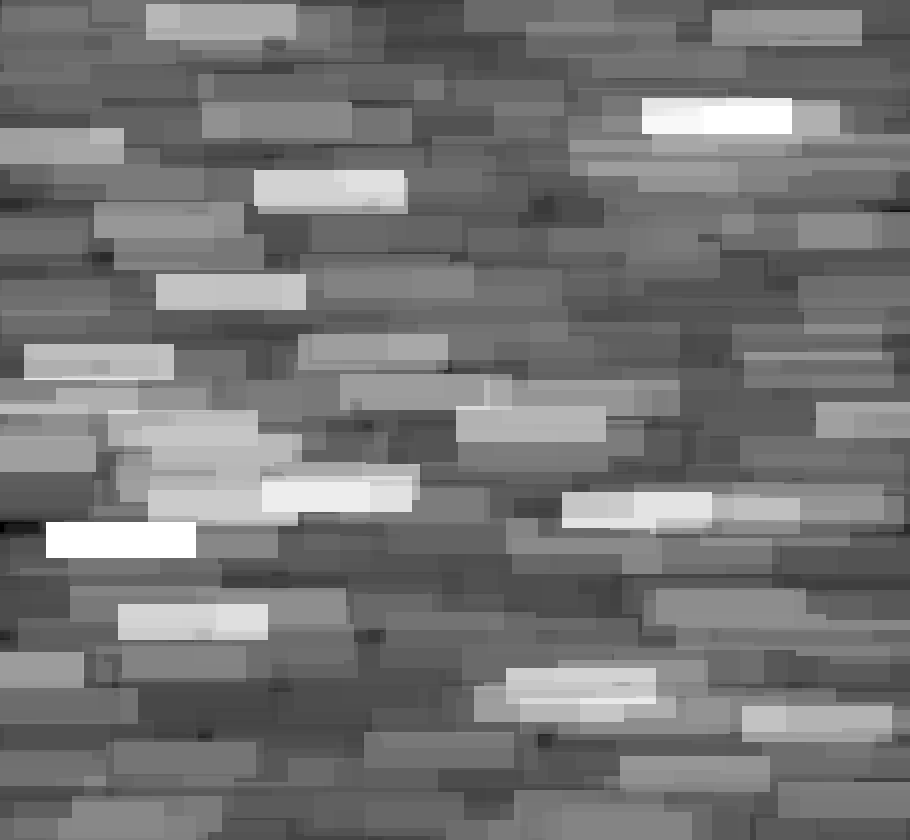}}
  \centerline{(d) Map $As_{\mathbf{t}}^{\protect \triangletimes}\tilde{\mathbf{f}}$}\medskip
\end{minipage}
\hfill
\begin{minipage}[b]{0.49\linewidth}
  \centering
  \centerline{\includegraphics[width=0.49\columnwidth]{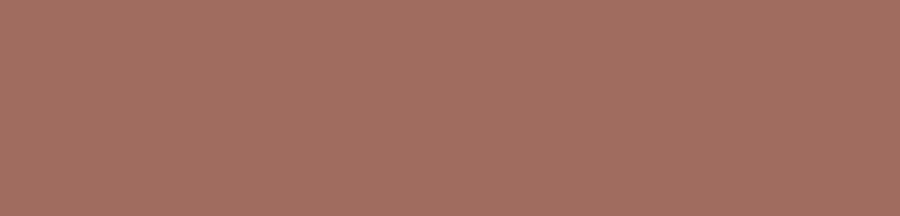}}
  \centerline{(e) Colour probe $\mathbf{t}$ magnified}
\end{minipage}
\begin{minipage}[b]{0.49\linewidth}
  \centering
  \centerline{\includegraphics[width=0.98\columnwidth]{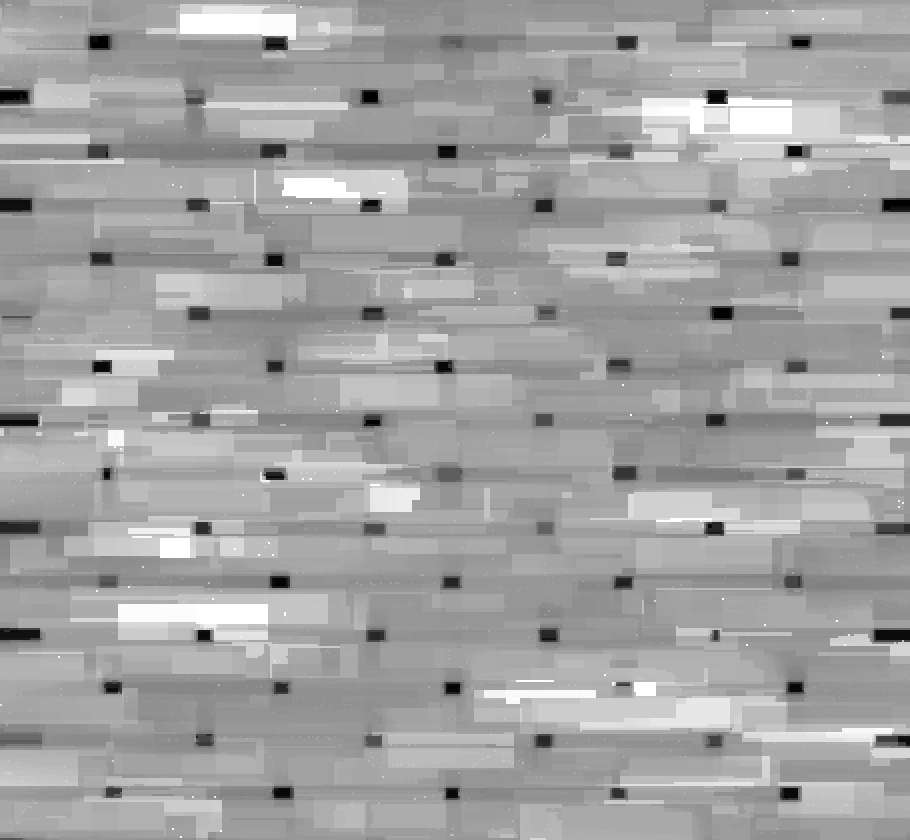}}
  \centerline{(f) Map $As_{\mathbf{t},p=98\%}^{\protect\triangletimes}\tilde{\mathbf{f}}$}
\end{minipage}
\hfill
\caption{Maps of Aspl\"und's distances without tolerance $As_{\mathbf{t}}^{\protect \triangletimes}\tilde{\mathbf{f}}$ and with tolerance $As_{\mathbf{t},p}^{\protect \triangletimes}\tilde{\mathbf{f}}$. In $\tilde{\mathbf{f}}$, a Gaussian white noise with zero-mean, variance 2.6 and spatial density $1\%$ has been used.}
\label{fig:brick}
\end{figure}

In figure \ref{fig:brick}, our aim is to find bricks of homogeneous colour inside a colour image of a brick wall. In the map of Aspl\"und's distances $As_{\mathbf{t}}^{\triangletimes}\mathbf{f}$ of the image without noise $\mathbf{f}$, the regional minima correspond to the location where the probe is similar to the image (according to Aspl\"und's distance). In the noisy image $\tilde{\mathbf{f}}$, the map of Aspl\"und's distance is very sensitive to noise (fig. \ref{fig:brick} d). Therefore, it is necessary to introduce a tolerance in the map $As_{\mathbf{t},p}^{\triangletimes}\tilde{\mathbf{f}}$, to find the minima corresponding to the bricks.
One can notice that: 
\begin{itemize}
	\item  the minima are preserved into the map with a tolerance (fig. \ref{fig:brick} c) compared to the map without (fig. \ref{fig:brick} f)
	\item  the maps are insensitive to a vertical lighting drift.
\end{itemize}
The minima can be extracted using standard mathematical morphology operations \cite{Matheron1967,Serra1982}. 
 
\begin{figure}[!htb]
\begin{minipage}[b]{0.32\linewidth}
  \centering
  \centerline{\includegraphics[width=\columnwidth]{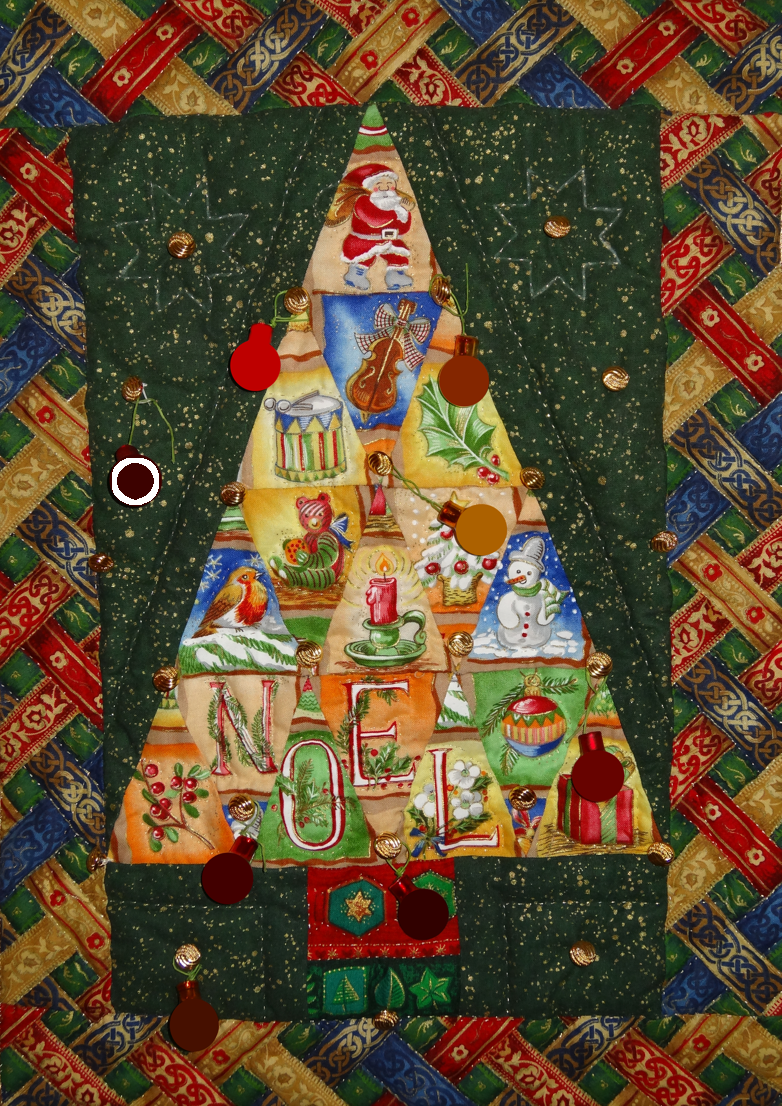}}
  \centerline{(a) Initial image $\mathbf{f}$}
	\centerline{and probe $\mathbf{t}$}
\end{minipage}
\begin{minipage}[b]{.32\linewidth}
  \centering
  \centerline{\includegraphics[width=\columnwidth]{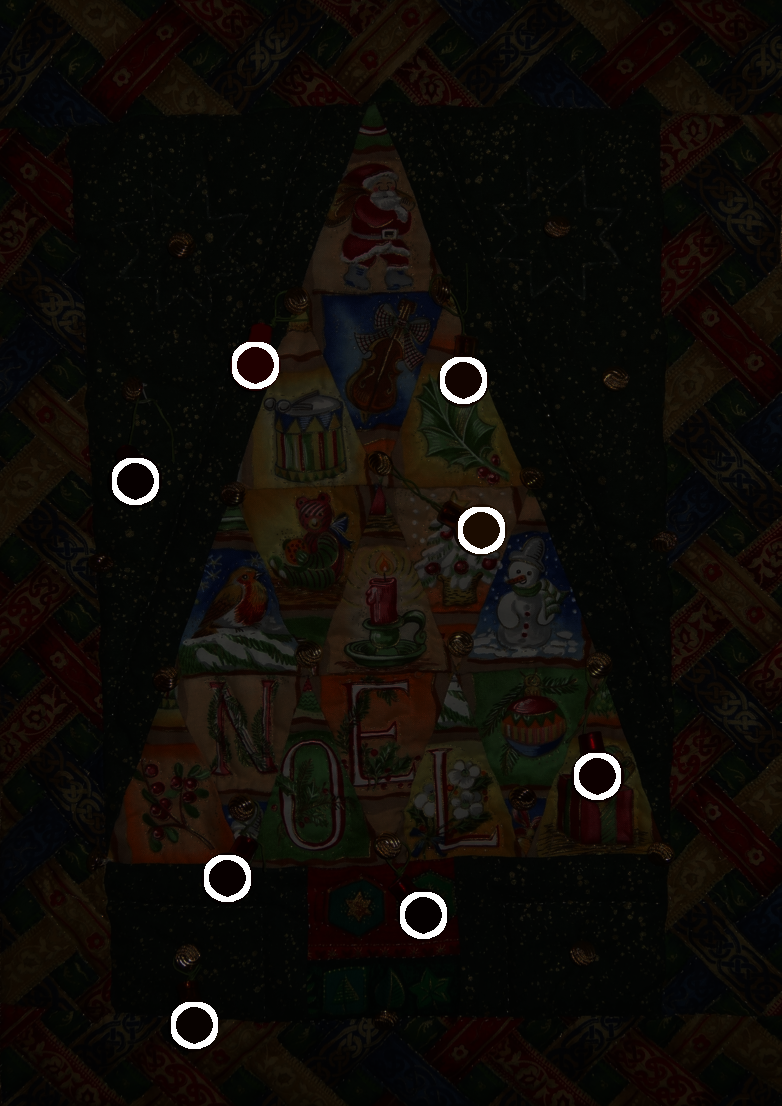}}
  \centerline{(b) Dark image $\tilde{\mathbf{f}}$}
	\centerline{Balls detected}
\end{minipage}
\begin{minipage}[b]{.32\linewidth}
  \centering
  \centerline{\includegraphics[width=\columnwidth]{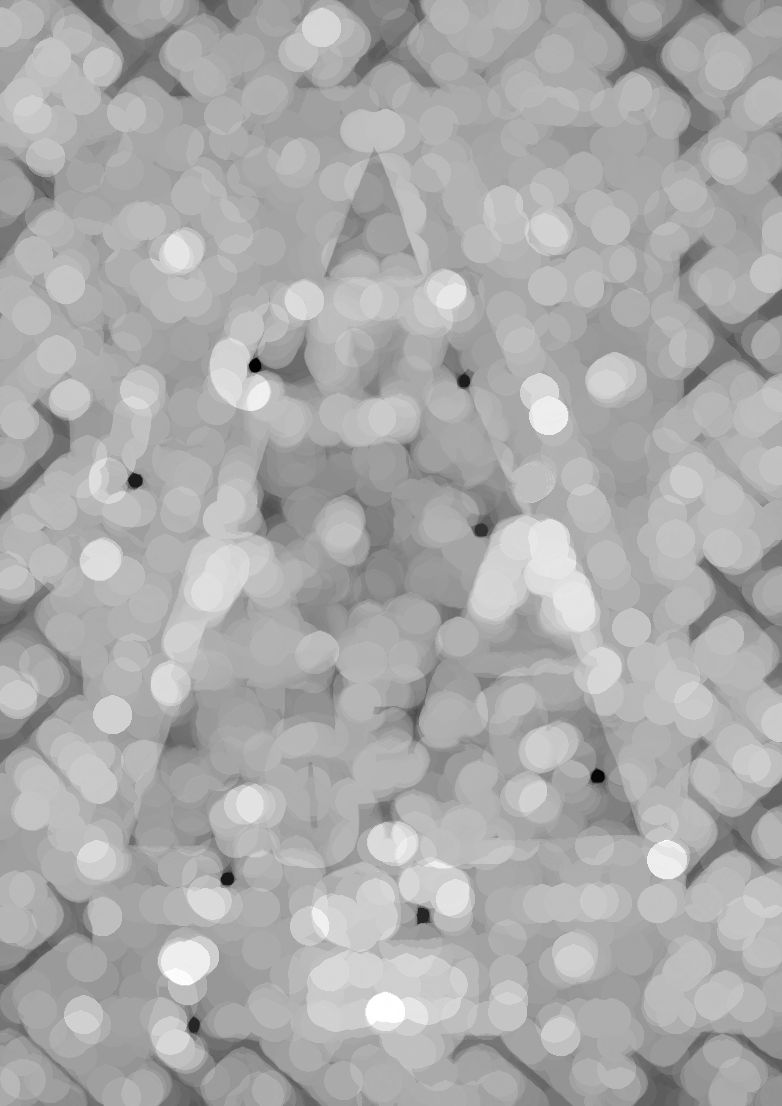}}
  \centerline{(c) Map $As_{\mathbf{t}}^{\triangletimes}\tilde{\mathbf{f}}$}
	\hfill
\end{minipage}

\caption{Detection of coloured balls on a dark image $\tilde{\mathbf{f}}$ with a probe $\mathbf{t}$ extracted in the bright image $\mathbf{f}$. (a) The border of the probe $\mathbf{t}$ is coloured in white.}
\label{fig:balls}
\end{figure}

In figure \ref{fig:balls}, two images of the same scene, a bright image $\mathbf{f}$ and a dark image $\tilde{\mathbf{f}}$, are acquired with two different exposure time. The probe $\mathbf{t}$ is extracted in the bright image and used to compute the map of Aspl\"und's distance $As_{\mathbf{t}}^{\triangletimes}\tilde{\mathbf{f}}$ in the darker image. By finding the minima of the map, the balls are detected and their contours are added to the image of figure \ref{fig:balls} (b). One can notice that the Aspl\"und's distance is very robust to lighting variations.

%
%

\section{Conclusion and perspectives}
\label{sec_concl}

An Aspl\"und's distance for colour and multivariate images has been introduced. Based on a double-sided probing of a function, this distance is particularly insensitive to the lighting variations. Moreover an alternative definition of the colour Aspl\"und's distance robust to noise has been introduced. An example illustrates the robustness of the method to the lighting variations and to the noise. With this new distance, efficient colour or multivariate pattern matching can be performed in images with all the properties described in \cite{Jourlin2014}. In future works, we plan to present the definition of a colour Aspl\"und's distance with a colour LIP model \cite{Jourlin2011} (already defined). We are also going to study the links between Aspl\"und's probing and mathematical morphology.

%
%

\bibliographystyle{IEEEbib}
\bibliography{refs}

\end{document}